\documentclass[letterpaper, 10 pt, journal, twoside]{ieeetran}

\usepackage{amsmath,amsfonts}

\usepackage{algorithm}
\usepackage{algpseudocode}
\usepackage{array} 
\usepackage{booktabs}
\usepackage{multirow}
\usepackage[caption=false,font=normalsize,labelfont=sf,textfont=sf]{subfig}
\usepackage{textcomp}
\usepackage{stfloats}
\usepackage{url}
\usepackage{verbatim}
\usepackage{graphicx}
\usepackage{cite}
\usepackage{todonotes}

\begin{document}

\title{Who to Trust in Multi-Robot Systems: Joint Robot and Object Hypothesis Trust Inference via Neural-Symbolic Cross-Validation}

\author{Ruiyang Wang, Hao-Lun Hsu, Haocheng Meng, Songqiao Xie and Miroslav Pajic}



\maketitle
\thispagestyle{empty}
\pagestyle{empty}
\begin{abstract}
Multi-robot systems (MRS) offer significant advantages over single-robot platforms by enabling information sharing and coordination for different tasks in complex environments. However, reliable cooperation critically depends on inferring trust over both information sources and their reported hypotheses, particularly under partial observability and potential adversarial interference. Existing approaches frequently depend on pre-established trust relationships or side-channel information, which may not be available or reliable in realistic MRS deployments. This work introduces \emph{NeST-Bayes}, a neural-symbolic framework that jointly infers robot and hypothesis trust using only inter-robot observation consistency, without requiring preassigned robot trust,  reputation information, or side-channel identity signals. \emph{NeST-Bayes} infers instantaneous trust evidence scores for robots and object hypotheses by reasoning over ego-centric relational graphs constructed from cross-robot observation consistency. These evidence scores are accumulated over time through Bayesian updates of Beta-distributed trust beliefs to maintain temporally consistent trust estimates. Extensive evaluations show that \emph{NeST-Bayes} achieves the highest mean object precision and robot recall across all benchmark settings, and the highest robot precision and object recall in most settings, compared to trust-based methods. Under aggressive optimized attacks, it achieves 98.8\%/93.8\% robot precision/recall and 98.3\%/95.4\% object precision/recall. In an occlusion-rich Webots warehouse, it maintains 92.5\%/92.0\% robot precision/recall and 97.3\%/85.0\% object precision/recall. This shows joint improvements in the reliability and coverage of retained information.
\end{abstract}

\IEEEpeerreviewmaketitle

\vspace{-5mm}
\section{Introduction}  
By enabling coordinated decision-making and information exchange among multiple robots, multi-robot systems (MRS) achieve performance gains over single-robot platforms in perception, exploration, coverage, and planning. Seminal works \cite{burgard2005coordinated, cortes2017coordinated} established the value of cooperation in distributed robot teams, while more recent efforts improved perceptual robustness through multi-view sensing \cite{zhou2022multi} and enhanced search efficiency via information sharing and coordinated planning~\cite{wang2025comres}. 

While multi-viewpoint aggregation overcomes individual sensing limitations, real-world MRS must operate under partial observability and sensing uncertainty \cite{xu2023bandit}, unreliable communication \cite{mannucci2019provably}, and restricted fields of view \cite{catellani2023distributed}. More critically, shared estimation and planning become vulnerable in contested environments and scenarios, where adversarial robots may inject/utilize misleading information; e.g., Sybil attacks distort collective objectives through false identities \cite{gil2017guaranteeing}, and Byzantine behaviors corrupt distributed diffusion and state estimation \cite{li2020resilient, mitra2019resilient}. Any mechanism aggregating inter-robot information should therefore infer trust over both the robots providing reports and the hypotheses they produce.

Existing trust inference approaches in MRS typically rely on assumptions that limit their applicability in realistic settings. Early works~\cite{burnett2011trust, yao2013matri, asiri2016iot} treat trust as a pre-established or reputation-based quantity, while more adaptive methods~\cite{pierson2016adaptive, cheng2021general} infer it from hand-designed models or specifications of nominal behavior. Recent resilient hypothesis-testing methods \cite{cavorsi2023dynamic, cavorsi2024exploiting} exploit side-channel information that is often unavailable. Complementary sensor-fusion approaches reject inconsistent measurements without explicitly estimating the reliability of their sources~\cite{tian2022kimera,bahrami2025multi}. Bayesian methods such as MATE~\cite{hallyburton2025security} already use observation consistency to jointly estimate agent and track trust, with each estimate informing the other. However, their mappings from consistency checks to trust evidence rely on hand-designed rules. 
A critical limitation is the lack of methods to learn how typed consistency and contradiction relations can be combined, 
to infer robot and object hypothesis trust under potential adversarial reporting. 

Consequently, in this work, we introduce \emph{NeST-Bayes}, a \underline{Ne}ural \underline{S}ymbolic \underline{T}rust inference framework that learns relational evidence to jointly estimate robot and object-hypothesis trust from cross-robot observation consistency, with \underline{Bayes}ian temporal evidence accumulation through updates to Beta-distributed trust beliefs. For each robot, we construct an ego-centric heterogeneous graph whose typed edges encode observation agreement and contradiction using geometric and field-of-view (FoV) checks. A triplet encoder initializes node representations from their symbolic relation profiles, and relation-specific graph attention propagates evidence across robot and hypothesis nodes to produce instantaneous trust~scores. 
These scores are accumulated as fractional pseudo-counts in Beta distributions to maintain temporal trust beliefs. 

\emph{NeST-Bayes} requires no preassigned robot trust or side-channel identity signals at inference time. Evaluations under noisy sensing, normal and optimized adversarial reporting, increased adversarial ratios, and occlusion-rich Webots simulations show that \emph{NeST-Bayes}
generally achieves the highest precision while also leading in recall among trust-based methods. Component ablation studies show that relational propagation improves recall while maintaining high precision.

In summary, the main contributions of this work include:
\begin{itemize}
    \item A neural-symbolic trust inference framework that combines symbolic relation profiles with heterogeneous graph attention to jointly infer robot and object-hypothesis evidence, integrating the learned scores with Bayesian temporal evidence accumulation.
    
    \item An extensive evaluation against \emph{No Trust}, \emph{Na\"ive Bayesian}, and \emph{MATE} baselines under sensing noise, normal and optimized adversarial reporting, increased adversarial ratios, and occlusion-rich Webots simulations, together with component ablations demonstrating how symbolic relation initialization and typed relational propagation contribute to precision and recall.
\end{itemize}

\section{Related Work}
\subsection{MRS Vulnerabilities}
MRS are sensitive to sensing noise and partial observability \cite{lauri2022partially}, communication uncertainty \cite{zhang2012iq, yan2012robotic}, and malicious interference \cite{bezzo2014attack}. False data injection can destabilize coordination and navigation \cite{yeke2025automated}, and adversarial perception attacks inject spurious measurements that disrupt cooperative decision-making \cite{bahrami2025multi}. Byzantine behaviors in distributed estimation and learning have also been studied \cite{deng2021investigation, li2022byzantine}, where even a few compromised robots degrade performance by injecting false positives or omitting valid observations. These results motivate principled trust mechanisms that operate under limited information and unreliable sensing, without assuming nominal-mode behavior.

\subsection{Trust in MRS}
Trust estimation in MRS remains an open problem. Existing approaches derive trust evidence from explicitly defined behaviors or task specifications \cite{cheng2021general, guo2023tip}, or exploit side-channel information that effectively fingerprints the robot identity \cite{cavorsi2023dynamic, cavorsi2024exploiting}. More recent methods \cite{hallyburton2025security, hallyburton2025trust, hallyburton2024bayesian} cross-validate observations across robots, but rely on hand-designed rules for trust updates that, as we show, can be brittle when adversaries are numerous or cross-robot validation is sparse. \emph{NeST-Bayes} instead learns how symbolic consistency and contradiction relations combine through heterogeneous message passing, producing joint robot and object-hypothesis evidence for temporal trust updates.

\subsection{Neural-Symbolic AI}
Neural-symbolic methods combine the interpretability of symbolic reasoning with the representational flexibility of neural networks \cite{besold2021neural, garcez2019neural}, and have proven effective for logical inference \cite{rocktaschel2017end}, probabilistic logic programming \cite{manhaeve2018deepproblog}, and relational reasoning \cite{santoro2017simple}. In robotics, they have succeeded in task planning \cite{Capitanelli2023AFF}, visual and relational reasoning \cite{liang2024visualpredicator, wang2025neuro}, and manipulation \cite{namasivayam2023learning, silverlearning}. Neural Logic Machines (NLMs) \cite{dongneural} learn
multi-step relational reasoning by composing symbolic predicates over a shared entity domain. Trust inference in MRS instead involves heterogeneous robot and object-hypothesis entities connected by typed observation and consistency relations. Relation-aware graph neural networks with heterogeneous message passing
\cite{velivckovic2017graph,wang2019heterogeneous,hu2020heterogeneous} provide a sparse substrate for this setting. \emph{NeST-Bayes} is conceptually
related to NLM-style relational reasoning, but operates over ego-centric heterogeneous graphs rather than using the explicit predicate operators of an NLM.

\section{Problem Formulation}
\label{sec:problem}
We consider a multi-robot system
$\mathcal{R}=\{r_i\}_{i=1}^{M}$ operating in a workspace
$\mathcal{W}\subset\mathbb{R}^{d}$ over a finite horizon
$t\in\{0,\ldots,T\}$. The environment contains a set of dynamic objects
$\mathcal{O}$, where each object $o\in\mathcal{O}$ is represented by its
position state $\mathbf{p}_o(t)\in\mathcal{W}$ and unique identifier
$id(o)\in\mathcal{I}_{\mathrm{gt}}$. Each robot $r_i$ has state
\begin{equation}
    \mathbf{s}_i(t)
    =
    \big(\mathbf{p}_i(t),\theta_i(t),R_i^{\mathrm{sen}},\Phi_i\big),
\end{equation}
where $\mathbf{p}_i(t)$ is its position, $\theta_i(t)$ is its heading,
and $(R_i^{\mathrm{sen}},\Phi_i)$ define its range- and angle-limited FoV, namely $A_i(t)$. The visible
ground-truth objects are
\begin{equation}
    \mathcal{O}_i(t)=
    \{o\in\mathcal{O}\mid \mathbf{p}_o(t)\in A_i(t)\}.
\end{equation}

\subsection{Noisy Sensing and Local Information}

Each robot $r_i$, whether legitimate or adversarial, first receives a nominal detector output $\mathcal{Z}^{\mathrm{nat}}_i(t)$. Let $\lambda^s_{\mathrm{FN},i}$ and $\lambda^s_{\mathrm{FP},i}$ denote its nominal sensor false-negative and false-positive
rates, respectively. For each visible object $o\in\mathcal{O}_i(t)$, the detector produces a noisy measurement $(id(o),\hat{\mathbf{p}}_{i,o}(t))$ with probability $1-\lambda^s_{\mathrm{FN},i}$, where $\hat{\mathbf{p}}_{i,o}(t)=\mathbf{p}_o(t)+ \boldsymbol{\epsilon}_{i,o}(t)$.
The detector may also produce persistent sensor artifacts
$\mathcal{Z}^s_i(t)$ with false-positive rate
$\lambda^s_{\mathrm{FP},i}$, giving
\begin{equation}
    \mathcal{Z}^{\mathrm{nat}}_i(t)
    =
    \mathcal{Z}^{\mathrm{gt}}_i(t)\cup
    \mathcal{Z}^{s}_i(t).
\end{equation}
Here $\mathcal{Z}^{\mathrm{gt}}_i(t)$ collects the true-positive detections of visible objects retained with probability $1-\lambda^s_{\mathrm{FN},i}$, and $\mathcal{Z}^{s}_i(t)$ collects nominal sensor false positives.

Robots communicate with neighbors $\mathcal{N}_i(t)=\{r_j\neq r_i\mid
\|\mathbf{p}_i(t)-\mathbf{p}_j(t)\|\le R_{\mathrm{com}}\}$ within the communication range $R_{\mathrm{com}}$. Each robot forms its reported track set and broadcasts
\begin{equation}
\mu_i(t)
=
\big(\mathbf{s}_i(t),\mathcal{Z}_i(t)\big),
\qquad
\mathcal{Z}_i(t)
=\{\tau_i^k(t)\}_{k=1}^{N_i(t)},
\end{equation}
where
\begin{equation}
\tau_i^k(t)
=
\big(id_i^k,\hat{\mathbf{p}}_i^k(t)\big)
\end{equation}
denotes the $k$-th local track reported by robot $r_i$, consisting of an object identifier and its estimated position. Here, $id_i^k$ denotes a persistent track label supplied by the
perception and association front end, rather than a ground-truth object identity available to the trust inference module. We assume that such labels can be matched across robot reports, while allowing
residual association errors, including missed, incorrect, or duplicate associations. A track is active at time $t$ if it is included in $\mathcal{Z}_i(t)$.

\subsection{Reporting Policies}
The robots are partitioned as
$\mathcal{R}=\mathcal{R}^{\mathrm{leg}}\cup\mathcal{R}^{\mathrm{adv}}$, where the robot type is unknown to the trust inference algorithm.
A legitimate robot does not manipulate its detector output:
\begin{equation}
    \mathcal{Z}_i(t)=\mathcal{Z}^{\mathrm{nat}}_i(t),
    \qquad r_i\in\mathcal{R}^{\mathrm{leg}}.
\end{equation}

Following the perception-level attacks of \cite{bahrami2025multi, hallyburton2025security}, adversarial robots
receive nominal outputs from the same sensing model but manipulate which
hypotheses they report, while their broadcast states $\mathbf{s}_i(t)$ remain authentic,
e.g., verifiable through inter-robot localization \cite{cheng2021general}. 

Let $\mathcal{F}$ denote the set of persistent adversarial false hypotheses, with
$|\mathcal{F}|\approx\lambda_{\mathrm{FP}}^{a}|\mathcal{O}|$ and identifiers disjoint from $\mathcal{I}_{\mathrm{gt}}$. Let
$\mathcal{F}_i(t)\subseteq\mathcal{F}$ be the false hypotheses in $r_i$'s FoV. Consequently, multiple adversarial robots may report and mutually support the same persistent false hypothesis. These induce adversarial false-positive track observations
$\mathcal{Z}^{\mathrm{advFP}}_i(t)
=\{(id(f),\hat{\mathbf{p}}_f(t))\mid f\in\mathcal{F}_i(t)\}$ while suppressed nominal detections form
$\mathcal{Z}^{\mathrm{advFN}}_i(t)\subseteq\mathcal{Z}^{\mathrm{nat}}_i(t)$.

In this work, we consider two standard adversarial policies. 
Under the \emph{normal adversarial policy}, robot $r_i$ randomly suppresses nominal detections with probability $\lambda_{\mathrm{FN}}^{a}$ and injects
visible persistent false hypotheses:
\begin{equation}
    \mathcal{Z}_i(t)\hspace{-2pt}=\hspace{-2pt}
    \left(
    \mathcal{Z}^{\mathrm{nat}}_i(t)\setminus
    \mathcal{Z}^{\mathrm{advFN}}_i(t)
    \right)
    \cup
    \mathcal{Z}^{\mathrm{advFP}}_i(t),
    \; r_i\in\mathcal{R}^{\mathrm{adv}}.
\end{equation}

Under the \emph{optimized adversarial policy}, 
$r_i$ selects reports~in
\begin{equation}
    \mathcal{C}_i(t)
    =
    \mathcal{Z}^{\mathrm{nat}}_i(t)
    \cup
    \mathcal{Z}^{\mathrm{advFP}}_i(t),
\end{equation}
where $\mathcal{Z}^{\mathrm{nat}}_i(t)$ contains nominal detector hypotheses and $\mathcal{Z}^{\mathrm{advFP}}_i(t)$ contains persistent adversarial false hypotheses visible to $r_i$. 
For each
$h\in\mathcal{C}_i(t)$, let $x_h\in\{0,1\}$ indicate whether $h$ is reported. Let $\mathcal{Y}_j(t)=\{id\mid (id,\hat{\mathbf{p}})\in\mathcal{Z}_j(t)\}$
denote the identifiers reported by neighbor $r_j$. Using current neighbor states and one-step delayed reports, $r_i$ computes
\begin{align*}
    n_i^{+}(h,t)
    &=
    \sum_{r_j\in\mathcal{N}_i(t)}
    \mathbf{1}[\hat{\mathbf{p}}_h(t)\in A_j(t)]
    \mathbf{1}[id(h)\in\mathcal{Y}_j(t-1)],\\
    n_i^{-}(h,t)
    &=
    \sum_{r_j\in\mathcal{N}_i(t)}
    \mathbf{1}[\hat{\mathbf{p}}_h(t)\in A_j(t)]
    \mathbf{1}[id(h)\notin\mathcal{Y}_j(t-1)] .
\end{align*}

The optimized adversary then solves
\begin{equation}
\begin{aligned}
    \max_{\{x_h\}_{h\in\mathcal{C}_i(t)}}\quad
    &
    \sum_{h\in\mathcal{Z}^{\mathrm{advFP}}_i(t)}
    x_h\big(\delta_{+}(1+n_i^{+}(h,t))-n_i^{-}(h,t)\big) + \\
    &
    \sum_{h\in\mathcal{Z}^{\mathrm{nat}}_i(t)}
    (1-x_h)\big(\delta_{-}(1+n_i^{-}(h,t))-n_i^{+}(h,t)\big) \\
    \mathrm{s.t.}\quad
    &
    \sum_{h\in\mathcal{Z}^{\mathrm{advFP}}_i(t)} x_h
    +
    \sum_{h\in\mathcal{Z}^{\mathrm{nat}}_i(t)}(1-x_h)
    \le B_i,\\
    &
    x_h\in\{0,1\},\qquad \forall h\in\mathcal{C}_i(t).
\end{aligned}
\label{eq:adv_obj}
\end{equation}
Here $\delta_{+}$ weights the benefit of promoting adversarial false
hypotheses, $\delta_{-}$ weights the benefit of suppressing nominal
hypotheses, and $B_i$ limits the number of manipulations per step for stealthiness. The reported set is
\begin{equation}
    \mathcal{Z}_i(t)=\{h\in\mathcal{C}_i(t)\mid x_h=1\} \qquad r_i\in\mathcal{R}^{\mathrm{adv}}.
\end{equation}

\subsection{Trust Inference Objective}
A central trust module estimates a robot trust score
$\gamma_i(t)\in[0,1]$ for each $r_i$, while each robot $r_i$ maintains track trust scores $\kappa_i^k(t)\in[0,1]$ for its local tracks
$\tau_i^k(t)$. These scores are inferred at deployment from cross-robot observation consistency, without requiring prior trust, reputation, or ground-truth labels at inference time.

Given threshold $\psi$, the inferred legitimate robot set is
\begin{equation}
    \widehat{\mathcal{R}}^{\mathrm{leg}}(T)
    =
    \{r_i\in\mathcal{R}\mid \gamma_i(T)\ge\psi\}.
    \label{eq:robot_trust}
\end{equation}
For evaluation, local tracks are grouped into fused hypotheses
$\widehat{\mathcal{O}}_{\mathrm{all}}$. For each fused hypothesis
$\bar{o}$, let $\mathcal{P}_{\bar{o}}$ be the set of reporting
robot-track pairs. Its credibility is
\begin{equation}
    \Gamma_{\bar{o}}(T)
    =
    \frac{
    \sum_{(i,k)\in\mathcal{P}_{\bar{o}}:\gamma_i(T)\ge\psi}
    \gamma_i(T)\kappa_i^k(T)}
    {
    \sum_{(i,k)\in\mathcal{P}_{\bar{o}}:\gamma_i(T)\ge\psi}
    \gamma_i(T)
    },
    \label{eq:object_trust}
\end{equation}
and zero if the denominator is zero. The credible object set is
\begin{equation}
    \widehat{\mathcal{O}}_{\mathrm{fused}}(T)
    =
    \{\bar{o}\in\widehat{\mathcal{O}}_{\mathrm{all}}
    \mid \Gamma_{\bar{o}}(T)\ge\psi\}.
\end{equation}

The objective is to identify legitimate robots and credible object hypotheses despite natural sensor artifacts and adversarial false hypotheses.

\section{NeST-Bayes: Neural-Symbolic Trust with Bayesian Evidence Accumulation}

\begin{figure*}[ht!]
    \centering
    \includegraphics[width=1.0\linewidth]{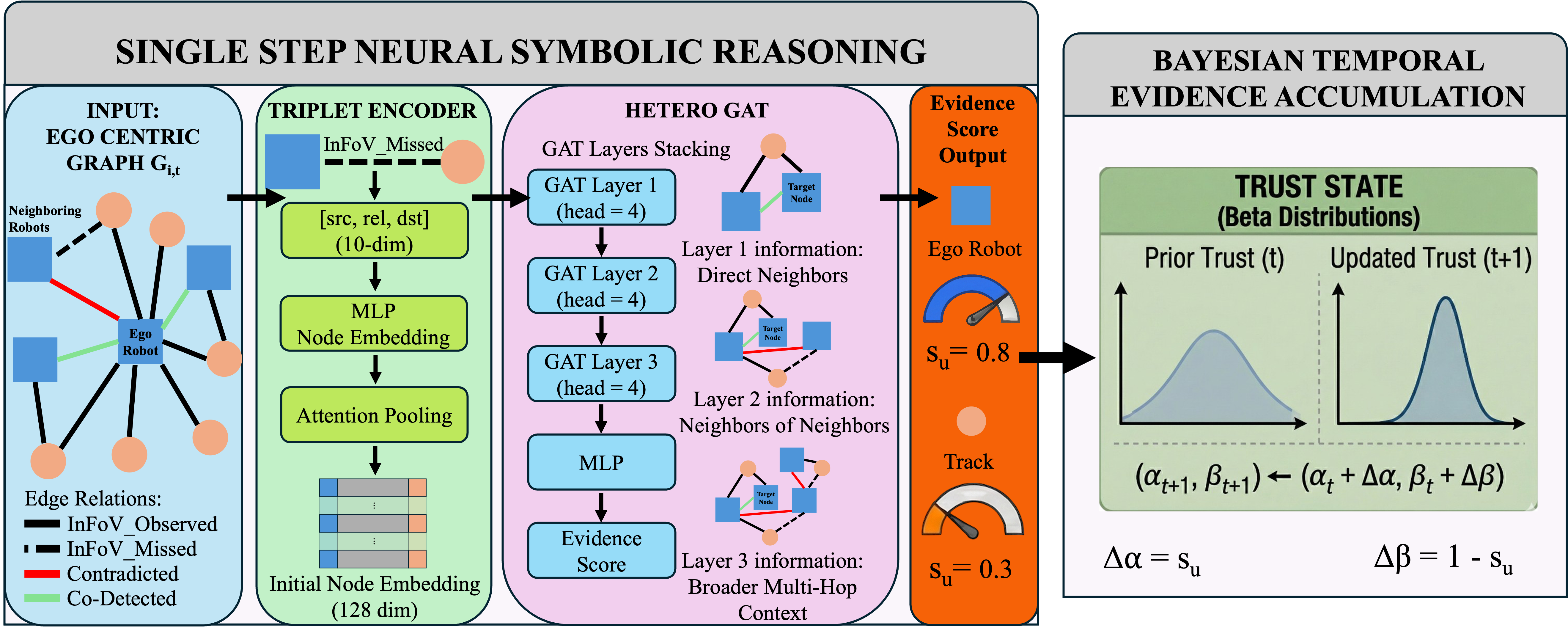}
    \caption{\small{\emph{NeST-Bayes} pipeline: ego-centric graph construction, neural-symbolic evidence inference, and Bayesian evidence accumulation.}}
    \label{fig:pipeline}
    \vspace{-4pt}
\end{figure*}

We introduce \emph{NeST-Bayes}, a trust inference framework based on ego-centric relational graphs encoding cross-robot observation consistency (Fig. \ref{fig:pipeline}). At each step, a centralized Trust Aggregation Module (TAM) constructs $\mathcal{G}_i(t)$ for each robot $r_i$ from $r_i$'s own and neighboring reports and applies a shared neural-symbolic model to reason through each graph, accumulating ego-robot evidence in a system-wide registry while returning validated track evidence to $r_i$, whose track beliefs are maintained onboard. The TAM could equally be replicated onboard each robot; we centralize it to obtain a single system-level trust estimate per robot, leaving the full decentralized deployment to future work.

\subsection{Single Step Neural-Symbolic Reasoning}
\label{sec:sing_step}

\subsubsection{Ego-Centric Graph Construction}
\label{sub:graph_construc}
At each time step $t$, the TAM constructs for robot $r_i$ a directed, heterogeneous ego-centric graph
$\mathcal{G}_i(t) = (\mathcal{V}_i(t), \mathcal{E}_i(t))$
from $r_i$'s message $\mu_i(t)$ and the messages
$\{\mu_j(t)\}_{r_j \in \mathcal{N}_i(t)}$ of its neighbors,
where each message contains the robot state and reported tracks. The graph translates the reported observations into a symbolic topology that facilitates trust inference.
\begin{equation}
\mathcal{V}_i(t)
=
\{r_i\}
\cup
\mathcal{N}_i(t)
\cup
\widetilde{\mathcal{T}}_i^{\mathrm{fused}}(t),
\end{equation}
where $\widetilde{\mathcal T}^{\mathrm{fused}}_i(t)$ is obtained by
grouping the tracks reported by $\{r_i\}\cup\mathcal N_i(t)$ according
to their persistent identifiers. Let $\mathcal I_i(t)$ denote the set
of identifiers appearing in these reports. For each
$id\in\mathcal I_i(t)$, define
\begin{equation*}
\begin{aligned}
\mathcal R_i^{id}(t)
&=
\left\{
r_j\in\{r_i\}\cup\mathcal N_i(t)
\,\middle|\,
\exists\,\tau_j^k(t)
=
\bigl(id,\hat{\mathbf p}_j^k(t)\bigr)
\in\mathcal Z_j(t)
\right\},\\
\widetilde{\tau}_i^{id}(t)
&=
\left(
id,\,
\frac{1}{|\mathcal R_i^{id}(t)|}
\sum_{r_j\in\mathcal R_i^{id}(t)}
\hat{\mathbf p}_j^{id}(t),\,
\mathcal R_i^{id}(t)
\right),
\end{aligned}
\label{eq:track_fusion}
\end{equation*}
where $\hat{\mathbf p}_j^{id}(t)$ is the position estimate associated
with $id$ in robot $r_j$'s report. The fused track set is
$\widetilde{\mathcal T}^{\mathrm{fused}}_i(t)
=
\{\widetilde{\tau}_i^{id}(t)\mid id\in\mathcal I_i(t)\}$, 
where $id$ is the persistent track label of Sec.~\ref{sec:problem}; residual missed, incorrect, or duplicate associations thus manifest as track-level false negatives or positives, so \emph{NeST-Bayes} operates on potentially imperfect associations.

\begin{table}[ht!]
    \centering
    \caption{Symbolic Edge Semantics in the Ego-Centric Graph}
    \label{tab:edge_semantics}
    \renewcommand{\arraystretch}{1.2}
    \begin{tabular}{p{1.3cm} p{2.1cm} p{4.0cm}}
        \toprule
        \textbf{Relation Category} & \textbf{Edge Type} & \textbf{Condition \& Interpretation} \\
        \midrule
        
        \multicolumn{3}{l}{\textit{\textbf{Robot–Track Relations (Sensing Consistency)}}} \\
        \hline
        
        \textbf{Observation} 
        & \texttt{InFoV\_Observed} 
        & A robot reports observing a fused track within its FoV. \\
        
        \textbf{Refutation} 
        & \texttt{InFoV\_Missed} 
        & A robot fails to observe a fused track within its FoV. \\
        
        \midrule
        
        \multicolumn{3}{l}{\textit{\textbf{Robot–Robot Relations (Cross-Robot Consistency)}}} \\
        \hline
        
        \textbf{Consistency} 
        & \texttt{Co-Detected} 
        & Both robots report observation of a fused track within their FoV. \\
        
        \textbf{Conflict} 
        & \texttt{Contradicted} 
        & A robot reports observation that another robot refutes within its FoV. \\
        \bottomrule
    \end{tabular}
\end{table}

The edges $\mathcal{E}_i(t)$ represent specific symbolic relations derived from geometric consistency checks and FoV analysis. These relations encode the logical ``evidence'' for or against a node's trustworthiness. To enable bidirectional information flow, in which robots validate tracks and tracks reciprocally validate robots, we model each relationship as a pair of directed edges (forward and reverse). This yields eight directed edge types, with the four forward relations defined in Table~\ref{tab:edge_semantics}. The TAM constructs this graph (Alg.~\ref{alg:single_step}, lines~1--7)
and then performs inference on it.

\begin{algorithm}[t]
\caption{Single Step Neural-Symbolic Reasoning}
\label{alg:single_step}
\begin{algorithmic}[1] 

\Require Neighbor messages $\mu_j(t)$ for all
$r_j\in\{r_i\}\cup\mathcal{N}_i(t)$

\Ensure Validated evidence scores
$\Omega_i=\{(u,s_u)\}$

\Statex \textbf{Phase 1: Ego-Centric Graph Construction}
\State $\mathcal{V}_i(t) \leftarrow     \{r_i\}
\cup
\mathcal{N}_i(t)
\cup
\widetilde{\mathcal{T}}_i^{\mathrm{fused}}(t)$, \; $\mathcal{E}_i(t) \leftarrow \emptyset$
\For{each $(u,v)\in\mathcal{V}_i(t)\times\mathcal{V}_i(t)$}
    \If{relation $\rho$ holds (Table~\ref{tab:edge_semantics})}
        \State $\mathcal{E}_i(t) \leftarrow \mathcal{E}_i(t)\cup\{(u,\rho,v),(v,\rho^{-1},u)\}$
    \EndIf
\EndFor

\Statex \textbf{Phase 2: Neural-Symbolic Reasoning}
\State $\mathcal{H} \leftarrow \text{ExtractTriplets}(\mathcal{E}_i(t))$
\State $\mathbf{h}^{(0)} \leftarrow \text{AttentionPooling}(\mathcal{H})$ 
\State $\mathbf{h}^{(1)}
\gets
\mathrm{Hetero GAT}^{(1)}
\bigl(\mathbf{h}^{(0)},\mathcal{E}_i(t)\bigr)$
\For{$\ell=2$ to $L$}
    \State $\hat{\mathbf{h}}^{(\ell)}
    \gets
    \mathrm{Hetero GAT}^{(\ell)}
    \bigl(\mathbf{h}^{(\ell-1)},\mathcal{E}_i(t)\bigr)$
    \State $\mathbf{h}^{(\ell)}
    \gets
    \hat{\mathbf{h}}^{(\ell)}
    +
    \mathbf{h}^{(\ell-1)}$
\EndFor

\Statex \textbf{Phase 3: Validity Filtering}
\State $\Omega_i\leftarrow\emptyset$
\For{each eligible node $u$}
    \If{IsValidNode$(u,\mathcal{E}_i(t))$}
        \State $s_u \leftarrow \sigma\big(\mathrm{MLP}_{\mathrm{type}(u)}(\mathbf{h}_u^{(L)})\big)$
        \State $\Omega_i\leftarrow\Omega_i\cup\{(u,s_u)\}$
    \EndIf
\EndFor
\State \Return $\Omega_i$ \Comment{Output evidence scores}

\end{algorithmic}
\end{algorithm}

\subsubsection{Neural-Symbolic Reasoning}
This stage maps the symbolic abstraction of Sec. \ref{sec:sing_step} into instantaneous evidence scores for both robots and object hypotheses, which are subsequently accumulated via Beta updates (Sec. \ref{sub:bayesian_update}). Rather than applying fixed aggregation rules to individual relations, we cast trust inference as typed relational reasoning over the ego-centric graph. Geometric and FoV consistency checks provide interpretable symbolic predicates, while stacked relation-specific message passing learns how local predicates combine into broader multi-hop evidence. This construction is inspired by neural-logic
relational reasoning, but does not instantiate the explicit predicate operators of an NLM.

In node feature initialization, as in the Triplet Encoder section in Fig.~\ref{fig:pipeline}, the semantic profile of a node is defined by the set of symbolic relations it participates in. For each node $u \in \mathcal{V}_i(t)$, we represent each relation from node $u$ to any other node $v$ as a symbolic triplet:
\begin{equation}
\eta(u,v) = [\phi(u), e({\rho}), \phi(v)] \in \{0,1\}^{10}
\label{eq:node}
\end{equation}
where $\phi(\cdot)$ is a one-bit node-type indicator (robot or track), and $e(\rho)$ is an 8-bit one-hot encoding of the relation type $\rho$ defined in Table~\ref{tab:edge_semantics} and their reverses. These triplets form an unordered bag of symbolic predicates, $\mathcal{H}_u$, that characterize the local consistency context of a node $u$.

Let $c=\mathrm{type}(u)\in\{\mathrm{robot},\mathrm{track}\}$. Each
triplet $\eta\in\mathcal H_u$ is embedded by a type-specific two-layer MLP $g_c$ and pooled according to \begin{equation}
\begin{aligned}
a_{\eta}
&=
\mathbf w_c^\top
\tanh\!\left(W_c g_c(\eta)\right),\qquad
\pi_{\eta}
=
\frac{\exp(a_{\eta})}
{\sum_{\eta'\in\mathcal H_u}
\exp(a_{\eta'})},\\
\mathbf h_u^{(0)}
&=
W_{o,c}
\sum_{\eta\in\mathcal H_u}
\pi_{\eta}g_c(\eta).
\end{aligned}
\label{eq:triplet_pooling}
\end{equation}
This permutation invariant pooling produces a $128$ dimensional initial embedding for the robot and track nodes.

The embeddings are processed by $L=3$ relation-specific Hetero GAT layers. Each layer applies standard graph attention independently over each directed relation type using $K=4$ heads. Head outputs and incoming relation representations are mean-aggregated, followed by ReLU and type-specific batch normalization. Residual connections are applied after the second and third layers, allowing evidence to propagate from local consistency relations to broader multi-hop consensus.

The final node embeddings, $\mathbf{h}_u^{(L)}$, are passed to type-specific MLP heads with a Sigmoid activation function $\sigma(\cdot)$ to produce scalar evidence scores $s_u \in [0,1]$.
\begin{equation}
    s_u = \sigma\big(\mathrm{MLP}_{\mathrm{type}(u)}(\mathbf{h}_u^{(L)})\big)
\end{equation}

\subsubsection{Validity Filtering}
\label{sec:validity}
The output of neural-symbolic reasoning is filtered before the trust update. Validity is defined differently for robots and tracks:
\begin{itemize}
    \item \textbf{Robots:} The ego robot $r_i$ is considered valid for a robot trust update only if it has at least one incident edge involving a neighboring robot in $\mathcal{N}_i(t)$.

    \item \textbf{Tracks:} A local track $\tau_i^k(t)\in\mathcal{Z}_i(t)$ is considered valid for a track trust update only if its corresponding fused track node is connected to the ego robot and at least one neighboring robot in $\mathcal{N}_i(t)$.
\end{itemize}

Although the GNN scores all nodes in $\mathcal{G}_i(t)$, only the valid ego-robot score and $r_i$'s local-track scores are retained. The TAM applies the ego score to $r_i$'s system-level trust belief, ensuring at most one robot update per step, and maps each valid fused-track score by its persistent identifier back to the local track $\tau_i^k(t)$ for $r_i$'s onboard update.

\subsection{Bayesian Temporal Evidence Accumulation}
\label{sub:bayesian_update}
To accumulate instantaneous neural evidence over time, we maintain a Bayesian trust belief for each node $u$, represented by a Beta distribution over its trustworthiness $\xi_u \in [0,1]$:
\begin{equation}
    f(\xi_u; \alpha_u(t), \beta_u(t)) = \frac{\xi_u^{\alpha_u(t)-1} (1-\xi_u)^{\beta_u(t)-1}}{\mathrm{Beta}(\alpha_u(t), \beta_u(t))}
\end{equation}
where $\alpha_u(t)$ represents the accumulated positive evidence and $\beta_u(t)$ represents the accumulated negative evidence up to time $t$. Accordingly, the TAM maintains $(\alpha_i,\beta_i)$ in a system-wide robot-trust registry, while each robot $r_i$ maintains $(\alpha_i^k,\beta_i^k)$ for its own local tracks. This division provides a globally consistent robot-trust estimate while keeping track beliefs colocated with the local perception and planning modules that consume them. Each newly encountered robot or track is initialized with the symmetric prior $(\alpha_0,\beta_0)=(1,1)$, corresponding to a neutral trust score of $0.5$. Entities receiving no valid cross-robot evidence retain this prior.

The Hetero GAT output score $s_u \in [0,1]$ is interpreted as the instantaneous evidence score. To ensure that evidence reinforces the correct belief parameter, we accumulate it as fractional pseudo-counts in the style of Beta-reputation and subjective-logic systems \cite{josang2002beta}:
\begin{equation}
    \alpha_u(t) = \alpha_u(t-1) + s_u, \quad \beta_u(t) = \beta_u(t-1) + (1 - s_u)
\label{eq:beta_update}
\end{equation}
This update is only applied to nodes that satisfy the validity constraints defined in Alg.~\ref{alg:single_step}. For non-valid nodes, the $\alpha$ and $\beta$ parameters remain unchanged.

The final trust score $\hat{y}_u(t)$ used for decision-making is the expected value of the Beta belief:
\begin{equation}
    \hat{y}_u(t) = \mathbb{E}[\xi_u] = \frac{\alpha_u(t)}{\alpha_u(t) + \beta_u(t)}
    \label{eq:expected_val}
\end{equation}
Over the horizon, the TAM repeats single-step reasoning for every robot, applies the update in Eq.~\eqref{eq:beta_update} to each validated evidence pair, and reports the final robot and track trust scores as the Beta expectations in
Eq. \eqref{eq:expected_val}, i.e., $\gamma_i(T)$ and $\kappa_i^k(T)$ in Eqs.~\eqref{eq:robot_trust}--\eqref{eq:object_trust}. This ensures trust reflects accumulated evidence rather than any single snapshot.

\section{Evaluations}

\begin{figure}
    \centering
    \includegraphics[width=0.6 \linewidth]{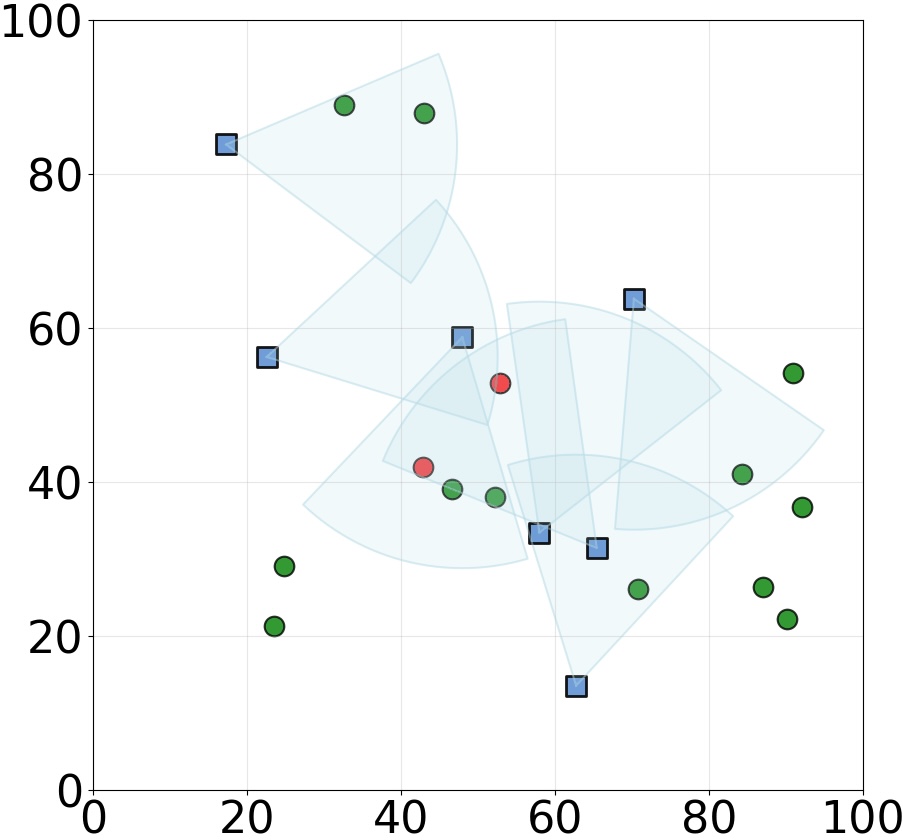}
    \caption{Example simulation environment. Robots are shown as blue squares with shaded fields of view, ground-truth objects as green points, and false positive objects as red points.}
    \label{fig:env}
\end{figure}

\begin{table*}[ht!]
\centering
\caption{Precision and recall under normal and optimized adversarial
policies. Each entry reports precision $\pm$ standard deviation
followed by recall $\pm$ standard deviation, in percent. Each setting
is evaluated over 100 independently generated scenarios.}
\label{tab:comprehensive_benchmark_results}
\renewcommand{\arraystretch}{0.90}
\scriptsize
\setlength{\tabcolsep}{2.0pt}

\begin{tabular*}{\textwidth}{
@{\extracolsep{\fill}}llcccc@{}
}
\toprule
\multirow{2}{*}{\textbf{Setting}} &
\multirow{2}{*}{\textbf{Entity}} &
\multicolumn{4}{c}{
\textbf{Precision $\pm$ Std.\,/\,Recall $\pm$ Std. (\%)}
} \\
\cmidrule(lr){3-6}
& &
\textbf{No Trust} &
\textbf{Na\"ive Bayesian} &
\textbf{MATE} &
\textbf{NeST-Bayes} \\
\midrule

\multicolumn{6}{l}{\textbf{Normal Adversarial Policy}} \\
\midrule

\multirow{2}{*}{\textbf{Nominal}}
& Robot
& $85.3{\pm}5.5/100.0{\pm}0.0$
& $93.3{\pm}8.8/79.2{\pm}20.6$
& $95.0{\pm}7.8/84.4{\pm}17.2$
& $\mathbf{98.6{\pm}3.7/95.8{\pm}8.2}$ \\
& Object
& $83.3{\pm}5.5/100.0{\pm}0.0$
& $94.7{\pm}7.4/95.1{\pm}10.9$
& $95.0{\pm}7.5/87.1{\pm}11.4$
& $\mathbf{98.0{\pm}5.0/93.7{\pm}8.4}$ \\
\midrule

\multirow{2}{*}{
\shortstack[l]{\textbf{High FP}\\
\textbf{Injection} $(\lambda_{\mathrm{FP}}^{a}\uparrow)$}
}
& Robot
& $85.3{\pm}5.5/100.0{\pm}0.0$
& $95.7{\pm}8.0/77.1{\pm}22.1$
& $95.3{\pm}7.5/84.3{\pm}17.2$
& $\mathbf{99.0{\pm}3.4/94.9{\pm}9.7}$ \\
& Object
& $67.1{\pm}4.3/100.0{\pm}0.0$
& $89.1{\pm}13.6/94.5{\pm}11.7$
& $90.9{\pm}13.7/86.7{\pm}11.1$
& $\mathbf{96.9{\pm}9.6/93.6{\pm}8.4}$ \\
\midrule

\multirow{2}{*}{
\shortstack[l]{\textbf{High FN}\\
\textbf{Suppression}
$(\lambda_{\mathrm{FN}}^{a}\uparrow)$}
}
& Robot
& $85.3{\pm}5.5/100.0{\pm}0.0$
& $97.4{\pm}14.2/75.5{\pm}25.1$
& $97.9{\pm}5.3/81.5{\pm}20.2$
& $\mathbf{99.4{\pm}2.2/94.3{\pm}10.7}$ \\
& Object
& $83.4{\pm}5.2/100.0{\pm}0.0$
& $95.5{\pm}7.3/91.5{\pm}16.2$
& $95.7{\pm}7.2/80.5{\pm}14.1$
& $\mathbf{99.1{\pm}3.3/93.3{\pm}8.7}$ \\

\midrule
\multicolumn{6}{l}{\textbf{Optimized Adversarial Policy}} \\
\midrule

\multirow{2}{*}{
\shortstack[l]{\textbf{Conservative} \textbf{Policy}}
}
& Robot
& $85.1{\pm}5.6/100.0{\pm}0.0$
& $88.8{\pm}8.1/86.9{\pm}17.0$
& $83.8{\pm}9.1/83.3{\pm}18.9$
& $\mathbf{86.5{\pm}7.6/92.5{\pm}11.4}$ \\
& Object
& $88.3{\pm}7.4/100.0{\pm}0.0$
& $95.5{\pm}5.9/94.5{\pm}12.1$
& $96.3{\pm}5.4/87.9{\pm}11.3$
& $\mathbf{97.0{\pm}5.1/94.4{\pm}7.4}$ \\
\midrule

\multirow{2}{*}{
\shortstack[l]{\textbf{Moderate}
\textbf{Policy}\\
$(\delta_{+},\delta_{-}\uparrow)$}
}
& Robot
& $85.1{\pm}5.6/100.0{\pm}0.0$
& $93.2{\pm}6.7/83.1{\pm}19.0$
& $90.8{\pm}9.0/80.2{\pm}18.3$
& $\mathbf{96.0{\pm}5.9/92.9{\pm}10.4}$ \\
& Object
& $87.1{\pm}7.2/100.0{\pm}0.0$
& $95.8{\pm}5.5/93.5{\pm}13.1$
& $97.8{\pm}4.6/83.5{\pm}13.5$
& $\mathbf{97.9{\pm}4.2/94.0{\pm}7.9}$ \\
\midrule

\multirow{2}{*}{
\shortstack[l]{\textbf{Aggressive}
\textbf{Policy} \\
$(\delta_{+},\delta_{-}\uparrow\uparrow)$}
}
& Robot
& $85.1{\pm}5.6/100.0{\pm}0.0$
& $96.5{\pm}5.3/79.9{\pm}21.0$
& $93.6{\pm}8.6/80.7{\pm}18.2$
& $\mathbf{98.8{\pm}4.2/93.8{\pm}10.4}$ \\
& Object
& $86.1{\pm}6.9/100.0{\pm}0.0$
& $97.1{\pm}5.2/93.5{\pm}13.1$
& $97.6{\pm}4.5/83.7{\pm}13.5$
& $\mathbf{98.3{\pm}3.7/95.4{\pm}6.7}$ \\
\midrule

\multirow{2}{*}{
\shortstack[l]{\textbf{High Adv.}
\textbf{Ratio}\\ 
$(m\in[0.4,0.5])$}
}
& Robot
& $62.1{\pm}4.0/100.0{\pm}0.0$
& $73.8{\pm}34.9/49.3{\pm}30.4$
& $58.9{\pm}25.4/59.0{\pm}27.8$
& $\mathbf{89.1{\pm}15.2/85.9{\pm}16.2}$ \\
& Object
& $80.6{\pm}5.6/100.0{\pm}0.0$
& $86.2{\pm}21.2/82.2{\pm}23.8$
& $81.5{\pm}21.4/67.4{\pm}23.6$
& $\mathbf{93.3{\pm}8.3/86.5{\pm}13.3}$ \\
\bottomrule
\end{tabular*}
\vspace{-8pt}
\label{tab:results}
\end{table*}

We first train and validate \emph{NeST-Bayes} in a controlled $100\times100$ 2D environment (Fig. \ref{fig:env}). Each robot has an occlusion-free FoV of $\pi/3$ and $50$-unit range, with $80$-unit communication range; we fix $|\mathcal{O}|=2|\mathcal{R}|$ and sample $|\mathcal{R}|\in[5,20]$. Robots and objects follow random start--goal motions over $100$ steps. During both training and testing, nominal sensor false-positive and false-negative rates are independently sampled from $[0.01,0.05]$ for every robot, including
legitimate and adversarial robots. Training samples the adversarial ratio $m\in[0.1,0.3]$ and, for normal adversaries, $\lambda^a_{\mathrm{FP}},\lambda^a_{\mathrm{FN}} \in[0.1,0.3]$, with scenarios split evenly between normal and optimized aggressive policies. From $10{,}000$ scenarios we extract $71{,}940$ valid ego-centric graphs, partitioned $80/20$ by scenario into train/validation. The model is supervised only on valid nodes (Sec. \ref{sec:validity}) via:
\begin{equation}
\mathcal{L}_i = 5\!\sum_{r\in\mathcal{V}^{\mathrm{tr}}_{R,i}}\!\ell_{\mathrm{BCE}}(y_r,s_r) + \!\sum_{\tilde\tau\in\mathcal{V}^{\mathrm{tr}}_{T,i}}\!\ell_{\mathrm{BCE}}(y_{\tilde\tau},s_{\tilde\tau}),
\end{equation}
where $y=1$ marks legitimate robots and real hypotheses; the factor $5$ balances the single ego-robot label against multiple track labels, and the Beta updates add no learned parameters.

\subsection{Baselines and Evaluation Metrics}
In addition to a \emph{No Trust} baseline, which indiscriminately trusts all robots and therefore propagates adversarial observations, we also compare the proposed \emph{NeST-Bayes} method with two representative trust aggregation approaches commonly used in multi-robot systems: $\textit{(i)}$ \emph{Na\"ive Bayesian}, similar to \cite{cheng2021general}, that maintains independent trust estimates for robot and track nodes by directly accumulating positive and negative evidence over time to update the beta trust distribution: \texttt{Co-Detected} and \texttt{InFoV\_Observed} edges are treated as positive evidence for robot and track nodes respectively, while \texttt{Contradicted} and \texttt{InFoV\_Missed} edges are treated as negative evidence. Importantly, this baseline consumes exactly the same symbolic relations as \emph{NeST-Bayes} but weights them by fixed rules and treats each robot and track as an independent evidence accumulator. Any gap therefore reflects how the evidence is interpreted rather than the evidence itself; $\textit{(ii)}$ \emph{MATE} \cite{hallyburton2025security} is a rule-based trust estimation method that maps perception consistency checks into trust pseudomeasurements (PSMs) and updates trust scores for its underlying beta distributions through hand-designed aggregation rules. In all tests, trust is accumulated over a horizon of 100 time steps. Final classifications use the fixed
threshold $\psi=0.5$. We report precision and recall for both robot and object classification to separately measure the reliability of the retained information and
the coverage of legitimate robots and ground-truth objects. Precision is averaged over scenarios and scored 0 for both robots and objects when a method retains no robot.

\subsection{Generalization with Normal Adversaries}
We evaluate normal adversaries that consistently report visible false-positive hypotheses while suppressing nominal detections with probability $\lambda^a_{\mathrm{FN}}$. Each setting uses 100 scenarios; results are in Table~\ref{tab:comprehensive_benchmark_results}. Across all three settings, \emph{NeST-Bayes} achieves the highest mean robot precision
(98.6--99.4\%) and recall (94.3--95.8\%) among trust-based methods, together with the highest object precision (96.9--99.1\%). Under high false-positive injection ($\lambda^a_{\mathrm{FP}}\in[0.4,0.8]$), it achieves 96.9\%/93.6\% object precision/recall, compared with 89.1\%/94.5\% for \emph{Na\"ive Bayesian} and 90.9\%/86.7\% for \emph{MATE}. Under high false-negative suppression ($\lambda^a_{\mathrm{FN}}\in[0.4,0.8]$), \emph{NeST-Bayes} leads in both precision and recall for both entity types. The comparison with \emph{Na\"ive Bayesian}, which accumulates identical symbolic evidence, demonstrates the value of learned relational reasoning for both the reliability and coverage of retained information (Sec.~\ref{sec:ablation}).
 
\subsection{Generalization with Optimized Adversaries}
\label{sec:optimized}
We next evaluate adversaries that optimize both the timing and type of their manipulations via Eq.~\eqref{eq:adv_obj}, with $\delta_+=\delta_-\in\{1,2,3\}$ yielding conservative, moderate, and aggressive policies and the attack budget fixed at $B_i=5$. More aggressive policies are preferable when the adversary seeks to maximally disrupt neighbors' motion planning: false positives divert robots toward nonexistent obstacles, while suppressed detections conceal real hazards. Results are shown in Table~\ref{tab:results}.

The conservative policy selectively injects false hypotheses that neighbors already appear to support and suppresses nominal detections lacking neighbor corroboration, producing
few observable inconsistencies. Across trust-based methods, robot precision remains close to the 85.1\% \emph{No Trust} baseline (83.8--88.8\%), indicating limited robot-level discrimination in this setting. \emph{NeST-Bayes} achieves 86.5\% robot precision with 92.5\% recall. At the object level, it increases precision from 88.3\% for \emph{No Trust} to 97.0\%, while retaining 94.4\% recall. This is the highest mean object precision among trust-based methods, compared with 95.5\% for \emph{Na\"ive Bayesian} and 96.3\% for \emph{MATE}. Its object recall is comparable to \emph{Na\"ive Bayesian}'s 94.5\% and exceeds \emph{MATE}'s 87.9\%. These results show that object-hypothesis validation remains effective when robot-level discrimination is limited.

As $\delta_+$ and $\delta_-$ increase, attacks become more detectable through cross-robot disagreement. \emph{NeST-Bayes} robot precision rises from 86.5\% under conservative attacks to 98.8\% under aggressive attacks, a 12.3-percentage-point gain compared with 7.7 points
for \emph{Na\"ive Bayesian} (88.8\% to 96.5\%). Object precision increases from 97.0\% to 98.3\%. Under aggressive attacks, \emph{NeST-Bayes} also achieves the highest robot and object recall among trust-based methods (93.8\% and 95.4\%). With the adversarial ratio increased to $m\in[0.4,0.5]$, \emph{NeST-Bayes} leads in all four metrics: 89.1\% robot precision with 85.9\% recall and 93.3\% object precision with 86.5\% recall. Compared with \emph{Na\"ive Bayesian}, robot precision and recall improve by 15.3 and 36.6 percentage points, respectively. These joint gains demonstrate effective filtering and retention of legitimate information when adversaries are numerous.

\subsection{Ablation Studies}
\label{sec:ablation}
We evaluate each ablation over 100 scenarios, evenly split between normal and optimized adversaries, with the latter equally sampling conservative, moderate, and aggressive
policies. Table~\ref{tab:component_ablation_results} shows that the full model achieves the
highest mean robot precision and recall among learned variants (97.2\% and 95.5\%), together with the highest object precision (99.0\%) and 94.8\% object recall. Removing the Hetero GAT reduces the model to node-wise MLPs and lowers robot recall from 95.5\% to 75.4\% and
object recall from 94.8\% to 89.3\%. These 20.1 and 5.5 percentage gains demonstrate the contribution of cross-node evidence propagation. Compared with a Homogeneous GAT, which collapses all relation types into a single edge type, typed relations improve robot precision and recall by 10.4 and 16.8 percentage points, respectively, and object precision by 12.0 points, showing the value of distinguishing agreement from contradiction. Replacing triplet-based initialization with learned node-type embeddings lowers robot precision from 97.2\% to 95.8\% and recall from 95.5\% to 91.3\%, demonstrating the benefit of encoding local symbolic relation profiles. Together, these results support the complementary contributions of symbolic initialization and heterogeneous relational reasoning.

\begin{table}[t!]
\centering
\caption{Component ablation over 100 comprehensive scenarios. Results are reported as mean $\pm$ standard deviation.}
\renewcommand{\arraystretch}{0.95}
\label{tab:component_ablation_results}
\setlength{\tabcolsep}{2.0pt}
\begin{tabular}{lcc}
\toprule
\multirow{2}{*}{\textbf{Method}} &
\multicolumn{2}{c}{\textbf{Precision / Recall (\%)}} \\
& \textbf{Robot} & \textbf{Object} \\
\midrule

\emph{No Trust}
& $84.7{\pm}5.4 \,/\, 100.0{\pm}0.0$
& $86.2{\pm}5.4 \,/\, 100.0{\pm}0.0$ \\

\emph{No Hetero GAT}
& $95.0{\pm}11.7 \,/\, 75.4{\pm}26.1$
& $98.5{\pm}3.6 \,/\, 89.3{\pm}12.6$ \\

\emph{Homogeneous GAT}
& $86.8{\pm}12.7 \,/\, 78.7{\pm}23.2$
& $87.0{\pm}5.8 \,/\, 96.1{\pm}6.6$ \\

\emph{Type Embeddings Only}
& $95.8{\pm}7.1 \,/\, 91.3{\pm}10.7$
& $98.2{\pm}4.0 \,/\, 94.4{\pm}7.1$ \\

\midrule
\textbf{NeST-Bayes (Full)}
& $\mathbf{97.2{\pm}6.2 \,/\, 95.5{\pm}8.8}$
& $\mathbf{99.0{\pm}2.9 \,/\, 94.8{\pm}7.8}$ \\

\bottomrule
\end{tabular}
\end{table}

\subsection{High Fidelity Simulations}
\label{sec:webots}
\begin{figure}[!t]
    \centering
\includegraphics[width=0.68\linewidth]{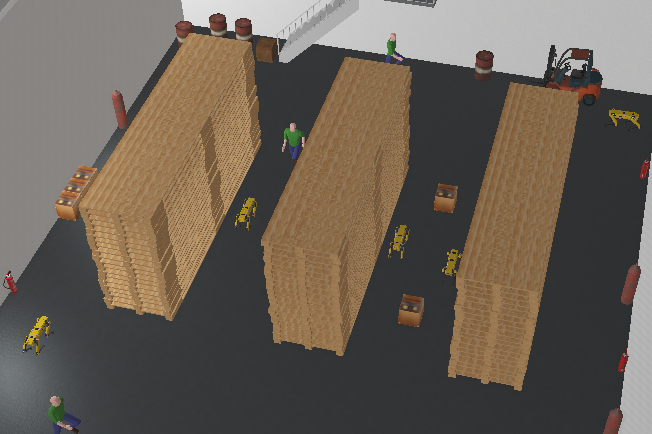}
    \caption{Webots warehouse simulation environment with five Spot robots and a mixture of objects. Occlusions from shelves and dynamic human agents create partial observability and asymmetric sensing among robots.}
    \label{fig:factory_world}
    \vspace{-4pt}
\end{figure}

We also evaluate \emph{NeST-Bayes} without retraining in a high-fidelity Webots~\cite{Webots} warehouse simulation with 5 Spot robots and 26 ground-truth objects, including dynamic (walking workers) and static objects (e.g., cardboard boxes falling from shelves), as in Fig.~\ref{fig:factory_world}. Each robot uses two head-mounted cameras ($\pi/4$ FoV, $20$\,m range), with FoV and occlusion reasoning from a known warehouse layout used only for visibility checks. Observations follow the same nominal sensing-noise model as in the previous sections. In each of 100 scenarios, two robots follow an optimized policy sampled from the conservative, moderate, and aggressive settings of Sec. \ref{sec:optimized}. The results are reported in Table~\ref{tab:webots}.

\begin{table}[t!]
\centering
\caption{Precision and recall in the Webots simulations over 100 scenarios. Results are reported as mean $\pm$ standard deviation.}
\vspace{-4pt}
\renewcommand{\arraystretch}{0.8}
\label{tab:webots}
\setlength{\tabcolsep}{1.0pt}
\begin{tabular}{lcc}
\toprule
\multirow{2}{*}{\textbf{Method}} &
\multicolumn{2}{c}{\textbf{Precision / Recall (\%)}} \\
& \textbf{Robot} & \textbf{Object} \\
\midrule

\emph{No Trust}
& $60.0{\pm}0.0 \,/\, 100.0{\pm}0.0$
& $71.4{\pm}6.3 \,/\, 100.0{\pm}0.0$ \\

\emph{Na\"ive Bayesian}
& $87.7{\pm}32.7 \,/\, 61.3{\pm}33.7$
& $87.7{\pm}32.7 \,/\, 81.6{\pm}31.5$ \\

\emph{MATE}
& $67.0{\pm}47.3 \,/\, 47.0{\pm}40.8$
& $67.0{\pm}47.3 \,/\, 38.1{\pm}28.8$ \\

\textbf{NeST-Bayes (Ours)}
& $\mathbf{92.5{\pm}13.1 \,/\, 92.0{\pm}14.3}$
& $\mathbf{97.3{\pm}5.4 \,/\, 85.0{\pm}7.5}$ \\

\bottomrule
\end{tabular}
\vspace{-8pt}
\end{table}

Frequent occlusions, limited FoV overlap, and dynamic objects create sparse cross-robot validation in Webots. \emph{NeST-Bayes} achieves the highest mean precision and recall for both entity types among trust-based methods: 92.5\% robot precision with 92.0\% recall, and 97.3\% object precision with 85.0\% recall. Relative to \emph{Na\"ive Bayesian}, robot precision and recall improve by 4.8 and 30.7 percentage points, respectively, while object precision and recall improve by 9.6 and 3.4 points. \emph{NeST-Bayes} also exhibits lower variability in robot precision (13.1 percentage points standard deviation) than \emph{Na\"ive Bayesian} (32.7) and \emph{MATE} (47.3). These results demonstrate effective transfer without retraining, improving both information reliability and coverage in an occlusion-rich~environment.

\section{Discussion and Conclusion}
\label{sec:limitations}
This paper presented \emph{NeST-Bayes}, a neural-symbolic framework that jointly infers robot and object-hypothesis trust from cross-robot observation consistency. By encoding observation agreement and contradiction as symbolic relations in ego-centric heterogeneous graphs and accumulating the resulting neural evidence through Bayesian updates of Beta-distributed trust beliefs, \emph{NeST-Bayes} operates without prior trust assignments, side-channel information, or hand-designed nominal behavior specifications. Comparisons against \emph{Na\"ive Bayesian}, which accumulates identical symbolic evidence, demonstrate the benefits
of learned relational reasoning for both precision and recall, including under occlusion-rich conditions. Component ablations show that heterogeneous relational propagation improves the coverage of legitimate robots and real objects. Across the benchmark settings, \emph{NeST-Bayes} achieves the highest mean object precision and robot recall among trust-based methods, while also leading in robot precision and object recall in most settings.

Future work will explore trust-aware active verification, directing robots to obtain additional viewpoints or revisit uncertain detections under sparse cross-robot evidence.
\bibliographystyle{IEEEtran}
\bibliography{bib}

\end{document}